\documentclass{article}

\usepackage[final]{galaxea}

\usepackage[utf8]{inputenc} 
\usepackage[T1]{fontenc}    
\usepackage{url}            
\usepackage{booktabs}       
\usepackage{amsfonts}       
\usepackage{graphicx}       
\usepackage{nicefrac}       
\usepackage{microtype}      
\usepackage{xcolor}         
\usepackage{amsmath, amssymb, bm}
\usepackage{algorithm}  
\usepackage[noend]{algpseudocode}  
\usepackage{algorithmicx}
\usepackage{array}
\usepackage{colortbl}
\usepackage{natbib}
\usepackage{caption}
\usepackage{hyperref}       
\usepackage{pifont}%
\usepackage{xspace}

\usepackage{graphics} %
\usepackage{epsfig} %
\usepackage{mathptmx} %
\usepackage{times} %
\usepackage{url}
\usepackage{array}  
\usepackage{multirow} 
\usepackage{makecell}

\usepackage{colortbl}
\usepackage{xcolor}
\usepackage{framed}
\usepackage{wrapfig}  
\usepackage{titletoc}
\usepackage{color} 
\usepackage{caption, subcaption, overpic, textpos}
\usepackage{titlesec}
\usepackage{comment}
\usepackage{fontawesome}
\usepackage{siunitx}
\usepackage{hyperref}      
\usepackage{tcolorbox} %
\tcbuselibrary{listings} %

\paperlogo{\includegraphics[height=0.6cm]{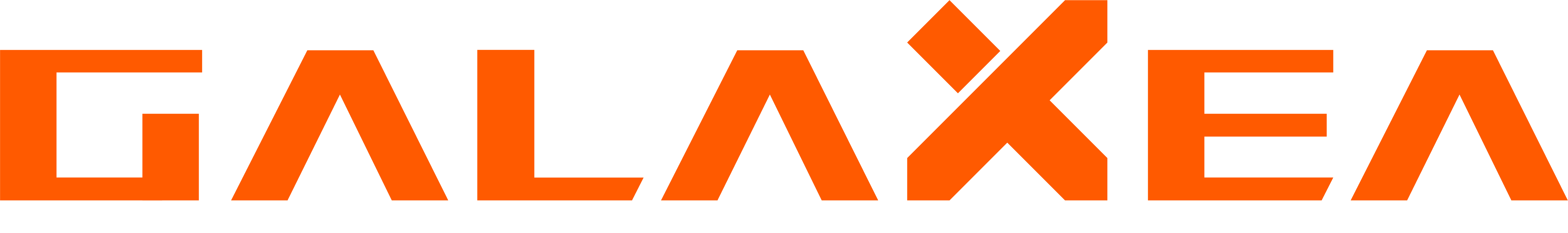}}

\usepackage{titlesec}
\definecolor{galaxeaOrange}{HTML}{FF5A00}

\titleformat{\section}
  {\sffamily\bfseries\Large\color{galaxeaOrange}}  
  {\thesection}{1em}{}  

\title{Galaxea Open-World Dataset and \\ G0 Dual-System VLA Model}

\author{%
  Galaxea Team \\
  \url{https://opengalaxea.github.io/G0/} \\
}

\begin{document}

\maketitle
\begin{abstract}

We present Galaxea Open-World Dataset, a large-scale, diverse collection of robot behaviors recorded in authentic human living and working environments. All demonstrations are gathered using a consistent robotic embodiment, paired with precise subtask-level language annotations to facilitate both training and evaluation.
Building on this dataset, we introduce G0, a dual-system framework that couples a Vision-Language Model (VLM) for multimodal planning with a Vision-Language-Action (VLA) model for fine-grained execution. G0 is trained using a three-stage curriculum:  cross-embodiment pre-training, single-embodiment pre-training, and task-specific post-training.
A comprehensive benchmark—spanning tabletop manipulation, few-shot learning, and long-horizon mobile manipulation—demonstrates the effectiveness of our approach. In particular, we find that the single-embodiment pre-training stage, together with the Galaxea Open-World Dataset, plays a critical role in achieving strong performance.

\end{abstract}

\begin{figure}[htbp]
        \centering
\includegraphics[width=1.0\textwidth]{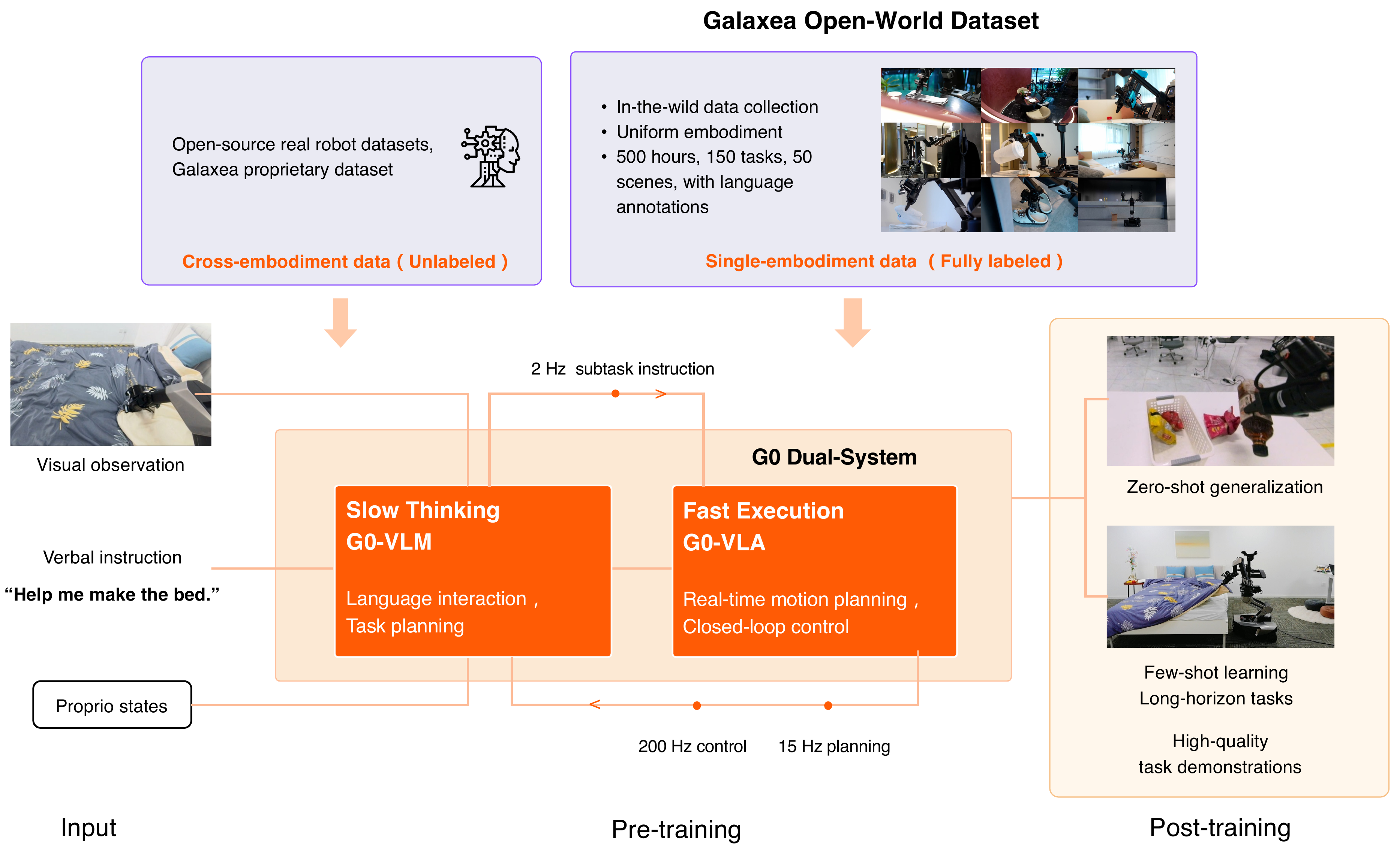} 
    \caption{We introduce Galaxea Open-World Dataset, a high-quality robot behavior dataset collected in the open world. Building on this dataset, we propose G0, a dual system which is composed of a VLM for slow thinking and a VLA model for fast execution.}
    \label{fig:teaser}
\end{figure}

\section{Introduction}
Vision-Language-Action (VLA) models have emerged as a pivotal paradigm aimed at enabling robots to autonomously perceive, reason, and perform complex tasks in the physical world. Despite significant progress, a substantial bottleneck persists due to the scarcity of large-scale, high-quality, open-world robot data.
Existing datasets, exemplified by Open-X Embodiment~\cite{vuong2023open}, are predominantly restricted by their limited task realism and insufficient environmental richness. These limitations impair the generalization of trained models when confronted with diverse real-world contexts.

In response to this challenge, we present \textbf{Galaxea Open-World Dataset}, an extensive, meticulously curated open real-world dataset for mobile manipulation. Galaxea Open-World Dataset comprises 500 hours of high-fidelity data systematically gathered in real-world scenarios where human individuals live and work, incorporating more than 150 distinct tasks across 50 different scenes. Uniquely, Galaxea Open-World Dataset was consistently captured using a single robotic embodiment, thereby ensuring uniformity and reliability. Comprehensive data filtering and precise language annotations further enrich the dataset, facilitating the benchmarking of mobile manipulation methodologies.

Complementing the dataset, we propose \textbf{G0, a dual system framework}. G0 capitalizes on System 2 (G0-VLM) for generalized multimodal planning, directing System 1 (G0-VLA) to perform precise action execution. 
These two models run asynchronously at different frequencies, enabling both efficient training and real-world deployment.
Importantly, we propose a 3-stage training curriculum for G0-VLA: (1) cross-embodiment pre-training on large-scale unlabeled datasets to acquire general world knowledge priors; (2) single-embodiment pre-training on our Galaxea Open-World Dataset to specialize in the perceptual-action pairs on the target platform; and (3) post-training on high-quality task demonstrations for the mastery of specific complex skills.

Finally, we construct a rigorous benchmark covering tasks such as tabletop manipulation, device operation, and long-horizon bed making in normal and few-shot learning settings. 
Experiments reveal that our high-quality dataset and the proposed pre-training strategy are effective in improving the dual system's performance.
Notably, when there is a large embodiment gap between the pre-training platform and the target robot, the benefits of cross-embodiment pre-training diminish or can even degrade the VLA model’s performance, underscoring the importance of the proposed single-embodiment pre-training stage.

The Galaxea Open-World Dataset and the models will be open-sourced in the coming weeks.
This release represents Galaxea team’s first step in open-sourcing robot datasets and models. 
We invite developers and researchers to build upon our work and push the boundaries of what is possible of embodied AI. 
\section{Related Work}

\textbf{Dual System Designs.} Our G0 model architecture builds on the foundation of hierarchical planning in robotics. 
In early methods such as Task and Motion Planning (TAMP)~\cite{garrett2021integrated}, high-level task planning and low-level motion control were often decoupled. 
The advent of VLMs has recently revitalized this paradigm. 
For example, SayCan~\cite{ahn2022can} demonstrated that a pretrained LLM can serve as a zero-shot planner for high-level goals. 
Inspired by this, the community has started to adopt dual-system frameworks based on Kahneman's theory of System 1 (fast, reactive) and System 2 (deliberative, planning)~\cite{kahneman2011thinking}. This hierarchical approach, separating deliberate planning from reactive control, forms the basis of our work.

\textbf{VLA as the System 1 Executor.} The rise of VLA models has provided a powerful paradigm for building generalist robot policies. 
Within a dual-system framework, these VLAs are a natural fit for the System 1 executor: a reactive policy that translates immediate sensory inputs and simple instructions into low-level robot control actions~\citep{chen2022visualgpt,shi2024lmfusion,wang2024cogvlm}. 

The action generation module in these VLA models employs two prevailing paradigms: autoregressive generation~\citep{szot2025multimodal,belkhale2024minivla,kim2024openvla,pertsch2025fast} and diffusion generation~\citep{black2024pi_0,liu2024rdt,li2024cogact}. Autoregressive models excel at transferring knowledge from pretrained VLMs but can be slow~\citep{wen2025tinyvla,pertsch2025fast}, while diffusion models offer higher throughput but risk degrading the VLM's original capabilities~\cite{intelligence2025pi_}. Currently, some work balances these two generation paradigms through more complex architectural designs~\citep{liu2025hybridvla}. 
Our work also implements a VLA for the System 1 component, employing two training methods similar to that of $\pi_{0.5}$~\citep{intelligence2025pi_} to leverage the benefits of both generation paradigms.

\textbf{VLM as the System 2 Planner.} While the System 1 VLA handles reactive control, it requires high-level guidance on what to do. This directive role is fulfilled by the System 2 planner. 
In modern robotic systems, this planner is typically implemented with a large VLM~\cite{cui2025openhelixshortsurveyempirical,shi2025hi,gao2025vla}. The VLM acts as the deliberative ``brain'', responsible for understanding a user's complex, open-ended command and decomposing it into a sequence of simpler sub-tasks, which are then passed to the System 1 executor.
The primary focus of our research is to systematically investigate this System 2 component. 
Specifically, we explore effective methods for constructing and fine-tuning the VLM planner. We conduct a systematic comparison using humans and closed-source models as baselines against an open-source VLM fine-tuned on our proprietary dataset, aiming to provide a practical reference for the community.

\textbf{Large-scale Manipulation Datasets.} The advancement of VLA models is fundamentally driven by large-scale, diverse, and high-fidelity robotic manipulation datasets. 
Current leading efforts have tackled this challenge along different axes. 
For instance, works like BridgeData V2~\cite{walke2023bridgedata} and DROID~\cite{khazatsky2024droid} focus on large-scale data collection on a specific robot embodiment. 
They successfully demonstrated the performance gains from scaling up data, but the single-platform setup inherently limits the diversity of their tasks and scenarios. 
In parallel, Open X-Embodiment~\cite{vuong2023open} significantly expanded embodiment and task diversity by aggregating data from numerous different sources. 
However, this heterogeneous aggregation introduced inconsistencies in data quality, annotation standards, and environmental context, creating potential noise for model training.

While these foundational datasets made crucial contributions to scale and diversity, they were predominantly collected in controlled or artificial settings. 
More recent efforts like RoboMIND~\cite{wu2024robomind} and AgiBot world~\cite{bu2025agibot}, despite pushing the boundaries of scale and task complexity, still operate within this limitation. 
This collection methodology results in a significant domain gap, impairing the ability of trained models to generalize to unstructured, real-world environments.
To directly address this challenge, we introduce a novel dataset uniquely characterized by its large-scale collection in completely unstructured, real-world settings. 
Our work aims to bridge this critical gap between existing data resources and real-world applications, providing a vital benchmark for training and evaluating truly robust VLAs.

Beyond building a more realistic dataset, we investigate how to leverage existing data best, focusing on the value of a common pre-training paradigm.
A widely adopted practice is to first pre-train a model on a large-scale, multi-embodiment dataset, and then continue training it on a target-embodiment-specific dataset. 
The goal is to synergize the benefits of broad generalization from diverse data with specialization from in-domain data. 
However, the true efficacy of this paradigm remains an open question with conflicting findings~\cite{black2024pi_0,zitkovich2023rt,zhang2024vlabench,shi2025diversity}. 
We argue that a key to resolving this is an unbiased, high-fidelity, real-world benchmark. 
Therefore, this work uses our novel dataset as a testbed. 
Through rigorous controlled experiments, we systematically analyze and disentangle the actual contribution of cross-embodiment pre-training to a model's generalization performance in the real world.

\section{Galaxea Open-World Dataset}

The Galaxea Open-World Dataset is a large-scale, high-quality, and fully annotated dataset drawn from Galaxea’s proprietary data collection efforts. 
It contains \textbf{100K demonstration trajectories} spanning \textbf{150 task categories} performed in \textbf{50 distinct real-world scenes}. 
These demonstrations cover more than \textbf{1,600 unique objects} and \textbf{58 operational skills}, ranging from fine-grained pick-and-place to coordinated whole-body manipulation. 
All data are collected with consistent embodiment, ensuring that perception, action, and language annotations are fully aligned across the dataset.

\begin{figure}[tbp]
    \centering
    \begin{subfigure}[b]{0.43\textwidth}
        \centering
        \includegraphics[width=\textwidth]{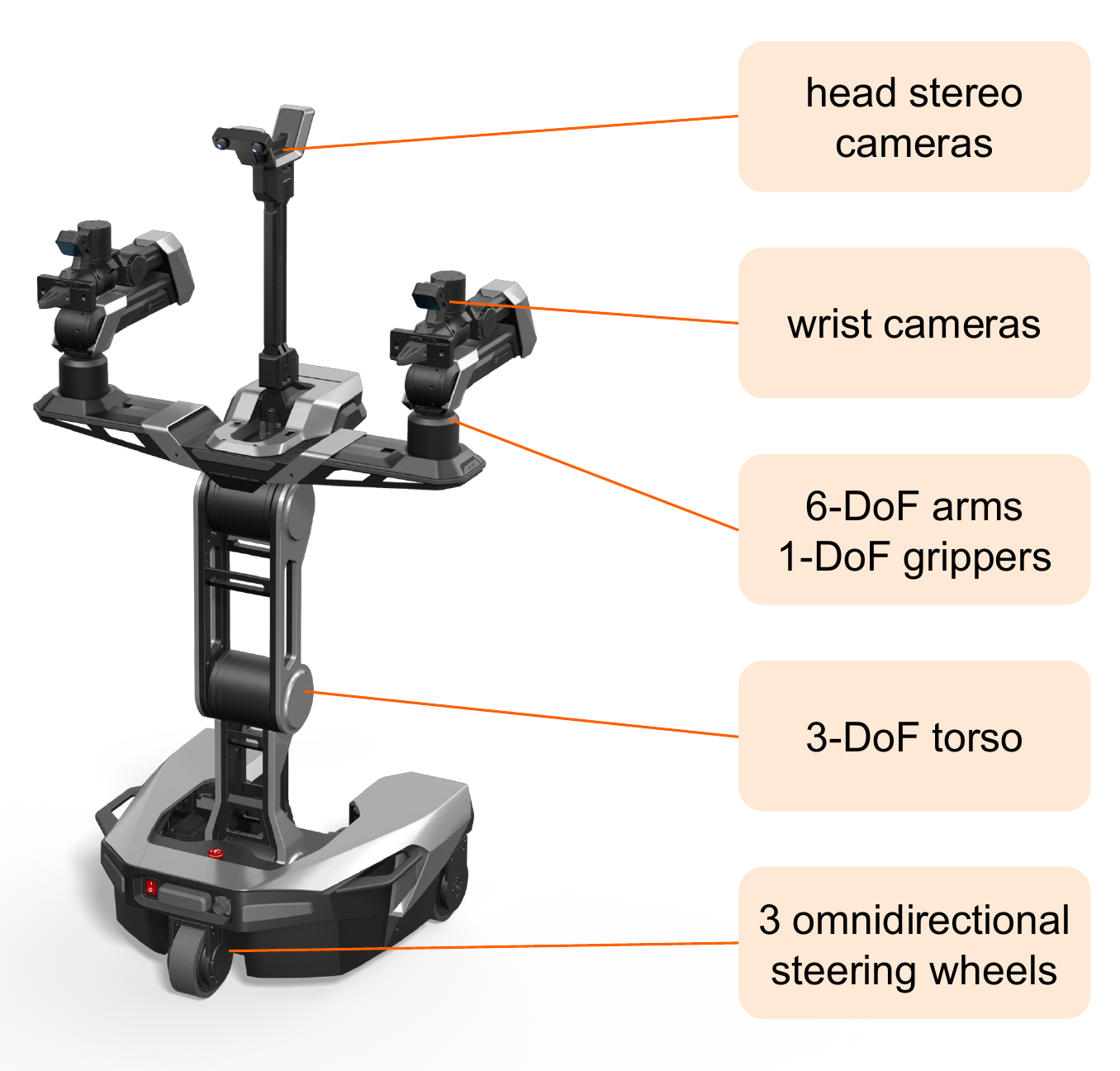}
        \caption{Galaxea R1 Lite platform.}
        \label{fig:r1lite}
    \end{subfigure}
    \hfill
    \begin{subfigure}[b]{0.53\textwidth}
        \centering
        \includegraphics[width=\textwidth]{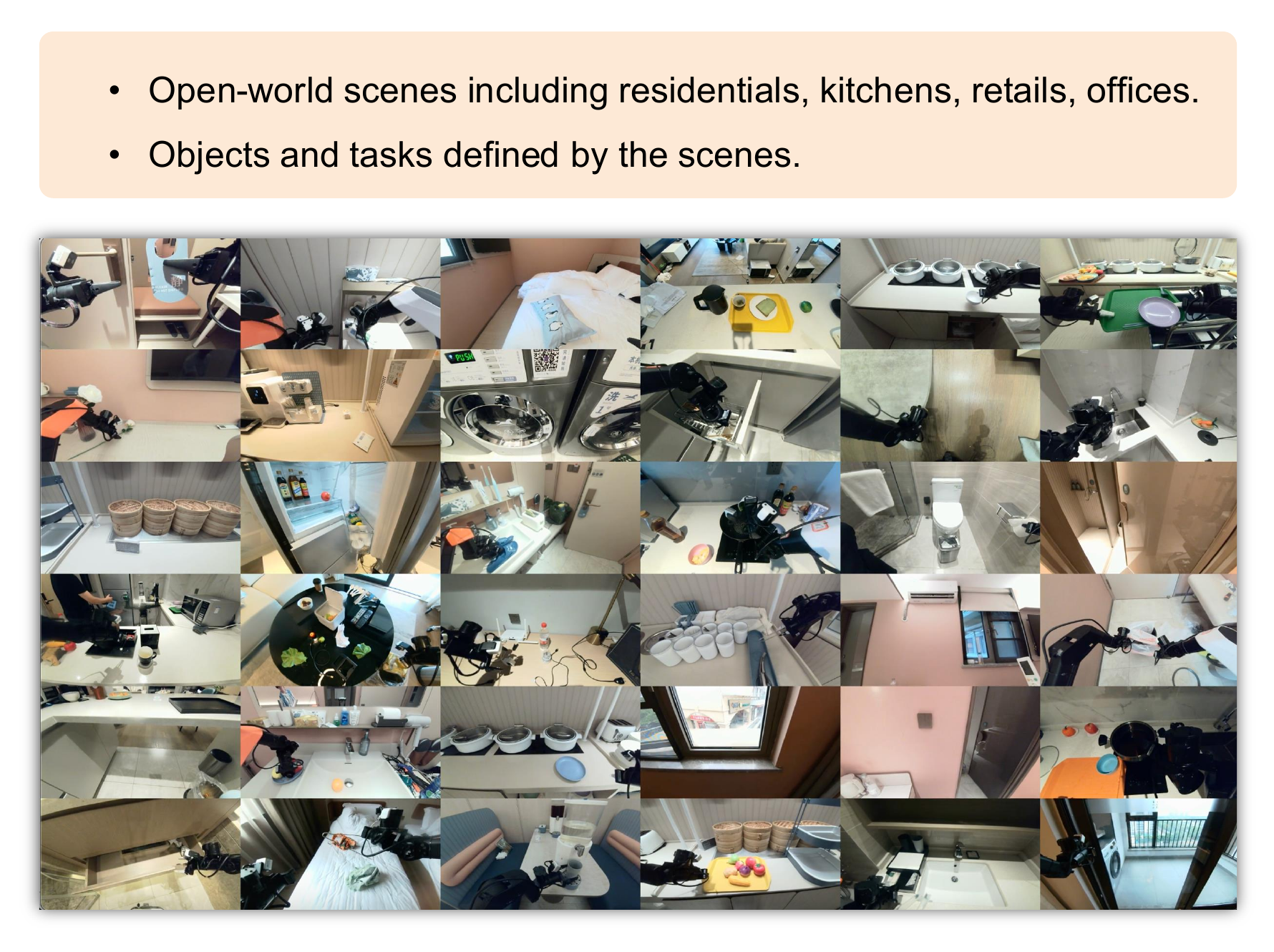}
        \caption{Data collected in diverse scenes.}
        \label{fig:data_samples}
    \end{subfigure}
    \caption{Galaxea Open-World Dataset is collected by a fleet of robots with identical embodiments, operating across diverse real-world environments.}
    \label{fig:data}
\end{figure}

\paragraph{Data Collection Platform.}
All demonstrations are recorded using the \textbf{Galaxea R1 Lite} platform (Fig.~\ref{fig:r1lite}), a mobile, bimanual robot designed for operation in human-centric environments. 
The robot features a 23-DoF embodiment comprising two 6-DoF arms, a 3-DoF torso with vertical and pitch motion for workspace extension, and a 6-DoF vector-drive omnidirectional base capable of speeds up to \SI{1.5}{m/s}. 
The spherical wrists and parallel grippers allow robust manipulation of everyday objects with payloads up to \SI{5}{kg} and reaches of \SI{60}{cm}.
Perception is provided by a stereo RGB head camera for scene-level context and dual Intel RealSense D405 RGB-D wrist cameras for close-range precision during manipulation. 
The robot’s compact dimensions (\SI{1280}{mm} height, \SI{600}{mm} chassis width, \SI{670}{mm} overall width) allow it to navigate through narrow spaces while maintaining manipulation capability.

To ensure natural and feasible motions, we adopt an \emph{isomorphic teleoperation} scheme that maps human operator movements directly to the robot’s kinematics. 
Compared to VR teleoperation, this approach keeps the arms within reachable postures, avoids inverse kinematics failures, and eliminates the need for re-targeting between human and robot morphologies.

\begin{figure}[tbp]
    \centering
    \begin{subfigure}[b]{0.3\textwidth}
        \centering
        \includegraphics[width=\textwidth]{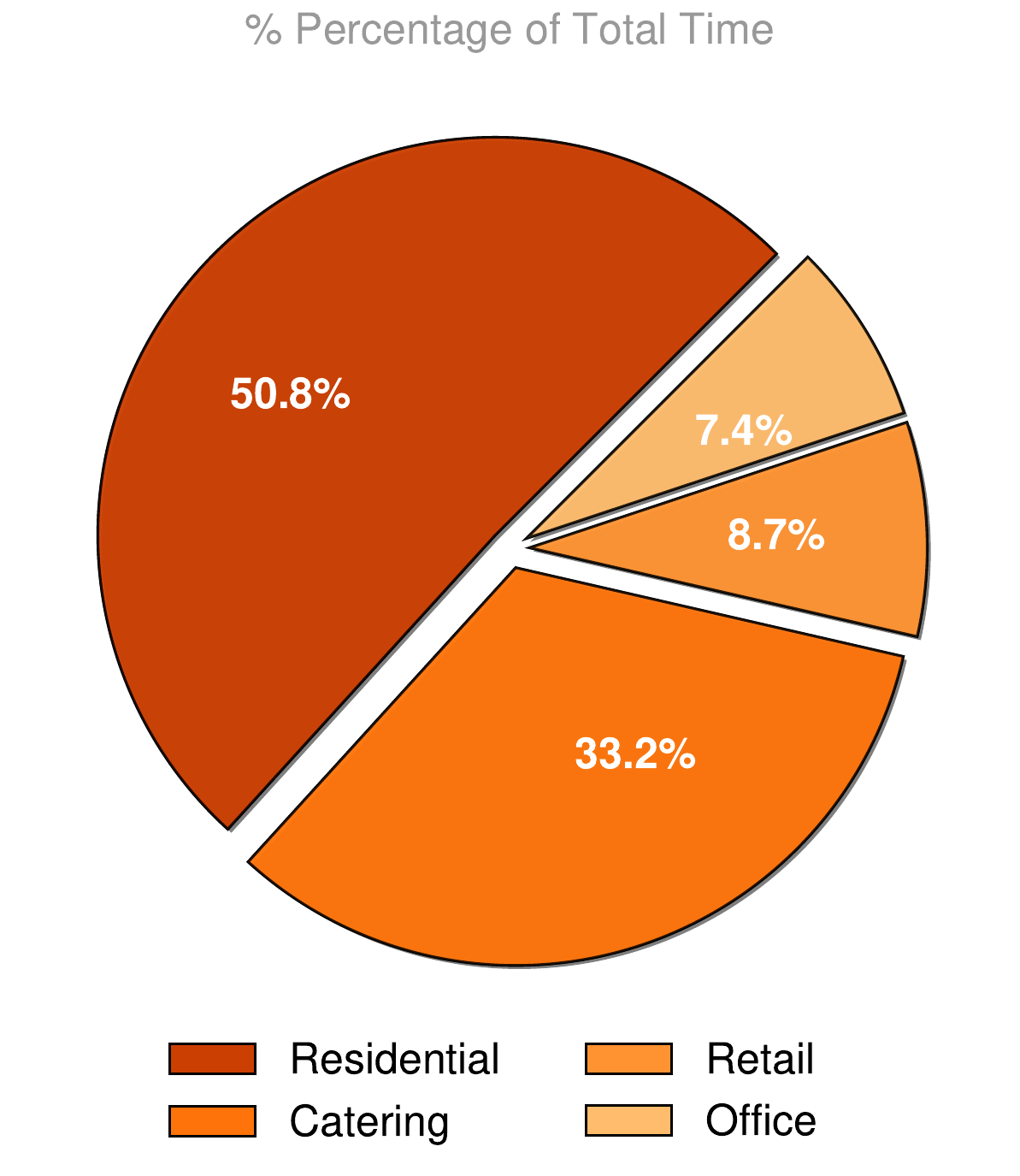}
        \caption{Scene distribution.}
        \label{fig:scene distribution}
    \end{subfigure}
    \hfill
    \begin{subfigure}[b]{0.65\textwidth}
        \centering
        \includegraphics[width=\textwidth]{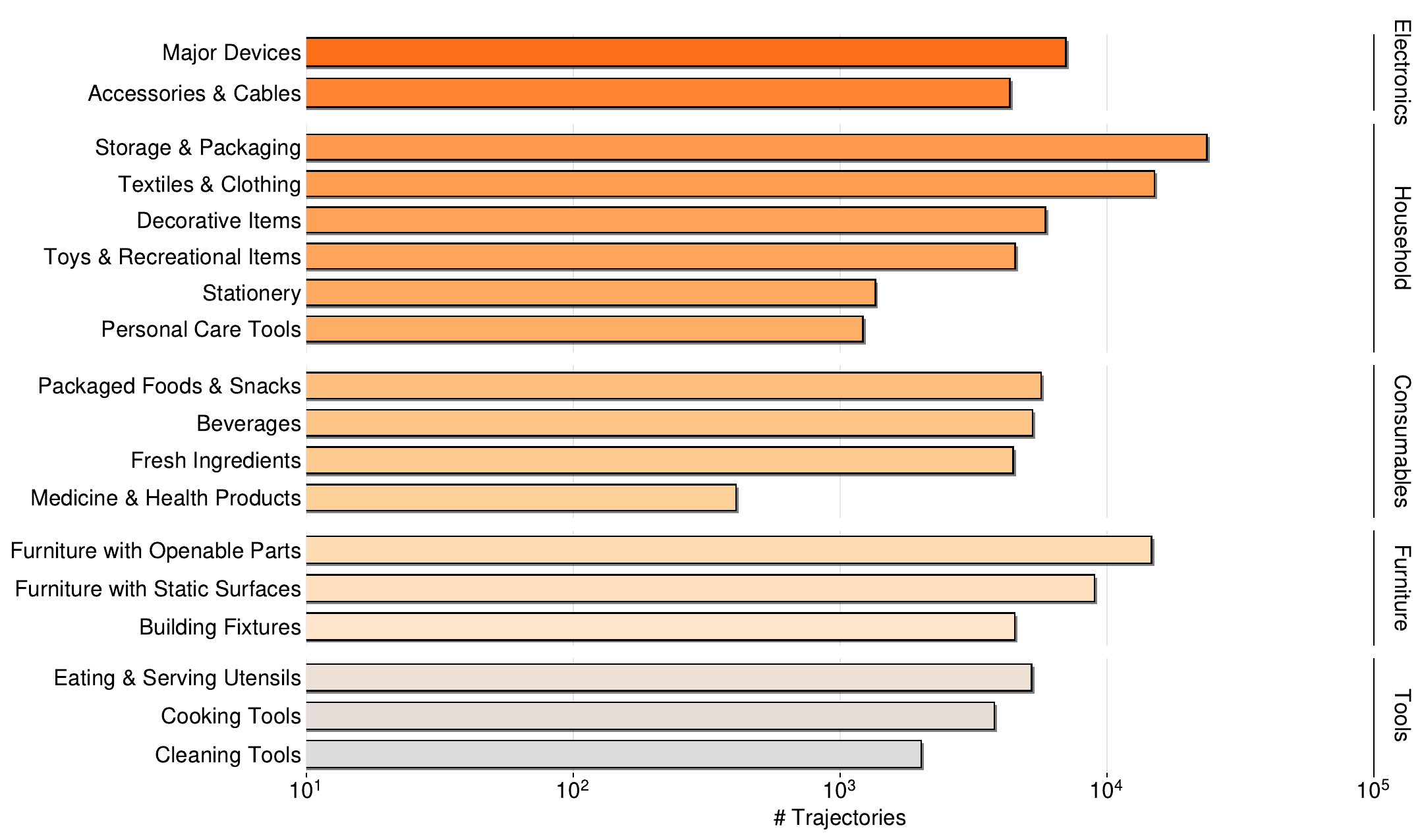}
        \caption{Object distribution.}
        \label{fig:object_distribution}
    \end{subfigure}
    
    \caption{\textbf{Data diversity statistics.} \textbf{(a)} The distribution of total interaction time is shown across the four primary scene categories: Residential, Retail, Catering, and Office. \textbf{(b)} Trajectory counts are presented for a rich collection of object subcategories, which are organized into broader classes like Electronics, Household, and Furniture, showcasing the dataset's wide range of interactive items.}
    \label{fig:data}
\end{figure}

\begin{figure}[tbp]
    \centering
    \begin{subfigure}[b]{0.45\textwidth}
        \centering
        \includegraphics[width=\textwidth]{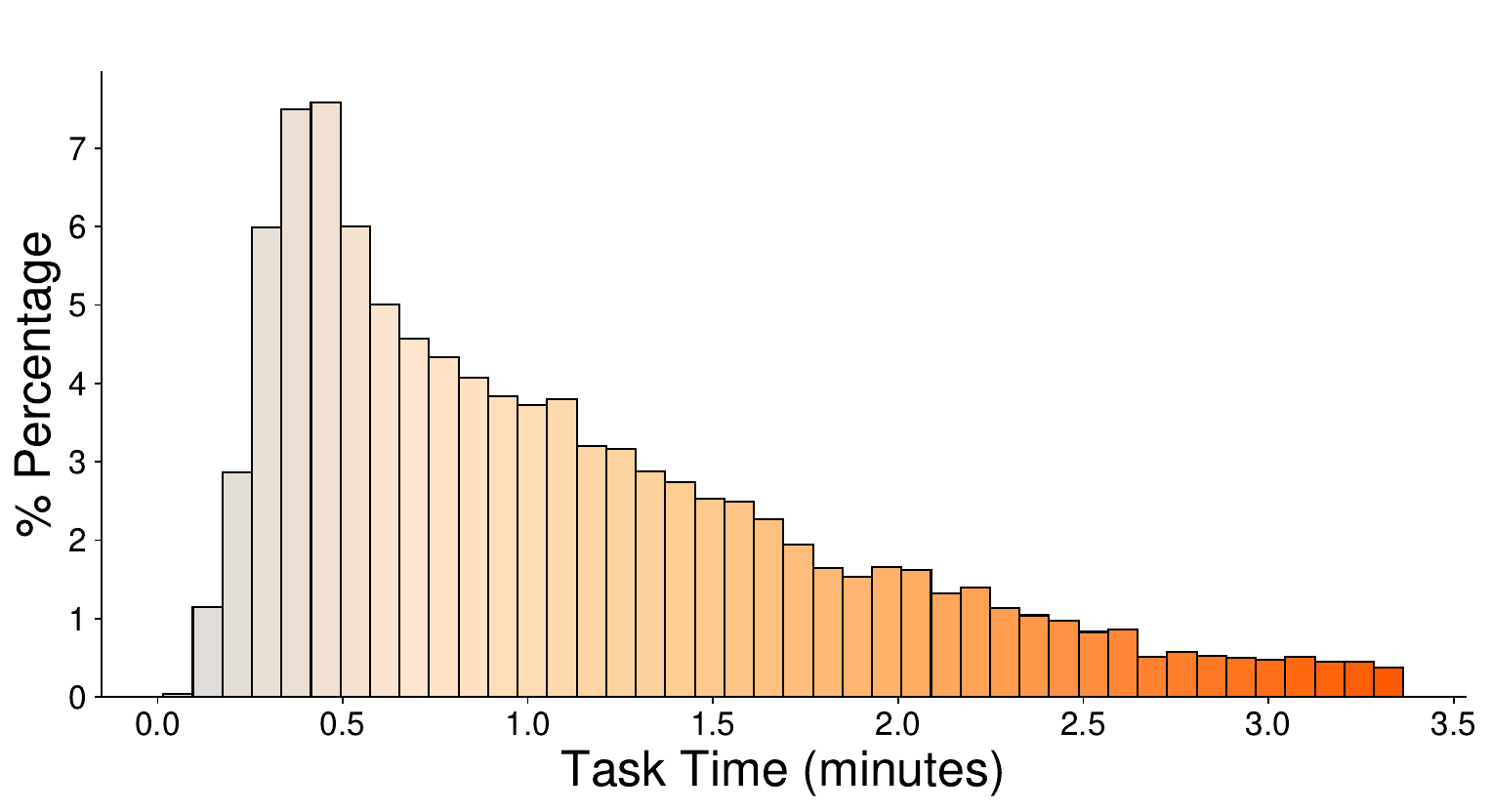}
        \caption{Task duration distribution.}
        \label{fig:task_duration_distribution} 
    \end{subfigure}
    \hfill
    \begin{subfigure}[b]{0.45\textwidth}
        \centering
        \includegraphics[width=\textwidth]{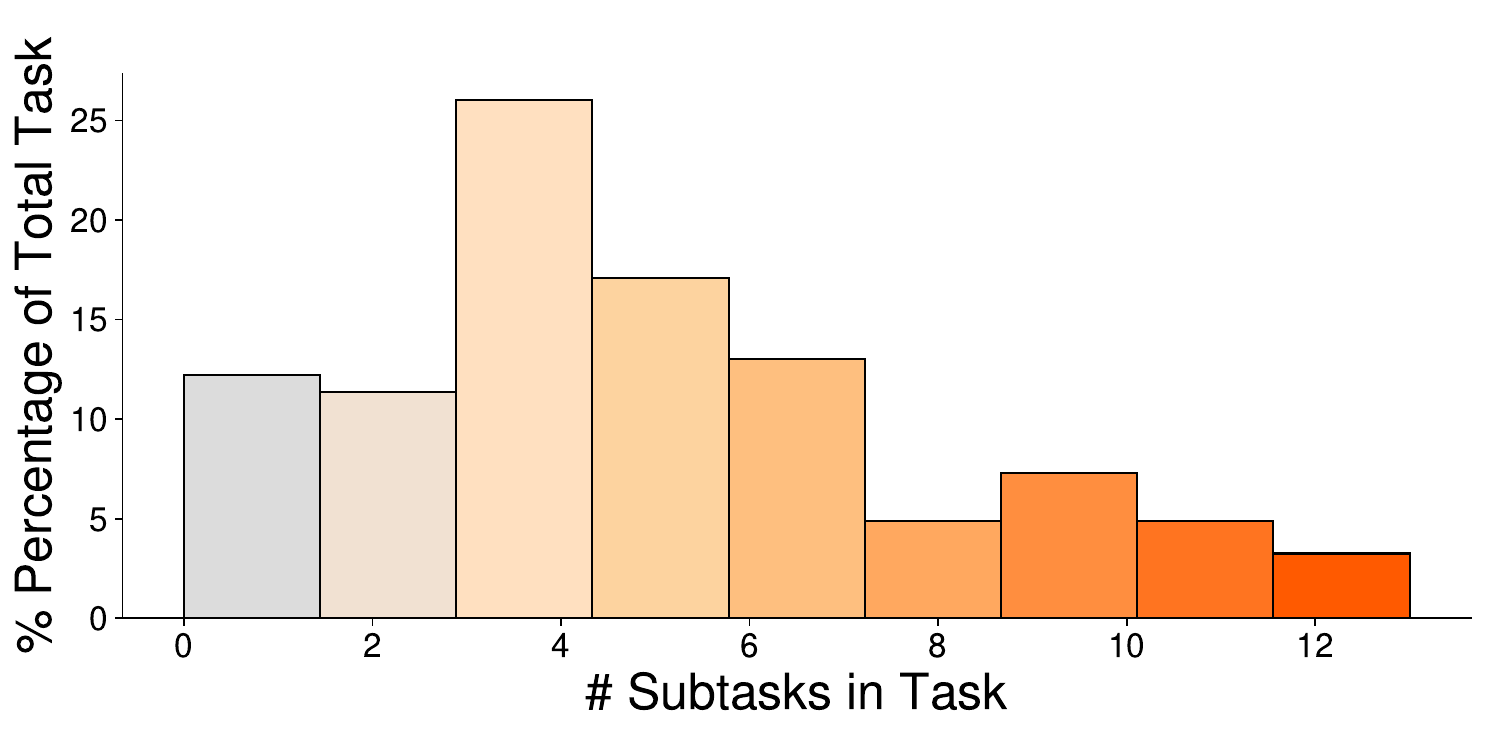}
        \caption{Task complexity distribution.}
        \label{fig:task_complexity_distribution} 
    \end{subfigure}
    
    \caption{\textbf{Task statistics.} This figure illustrates the temporal and structural properties of the tasks within the dataset. \textbf{(a)} The distribution of task completion times reveals that most tasks are of moderate length, yet the dataset also contains a long tail of complex, long-horizon activities. \textbf{(b)} Task complexity, measured by the number of subtasks per task, is shown to vary widely, covering everything from simple actions to intricate multi-step procedures.}
    \label{fig:task_stats}
\end{figure}

\begin{figure}[tbp]
    \centering

    \begin{subfigure}[b]{0.19\textwidth}
        \centering
        \includegraphics[width=\textwidth]{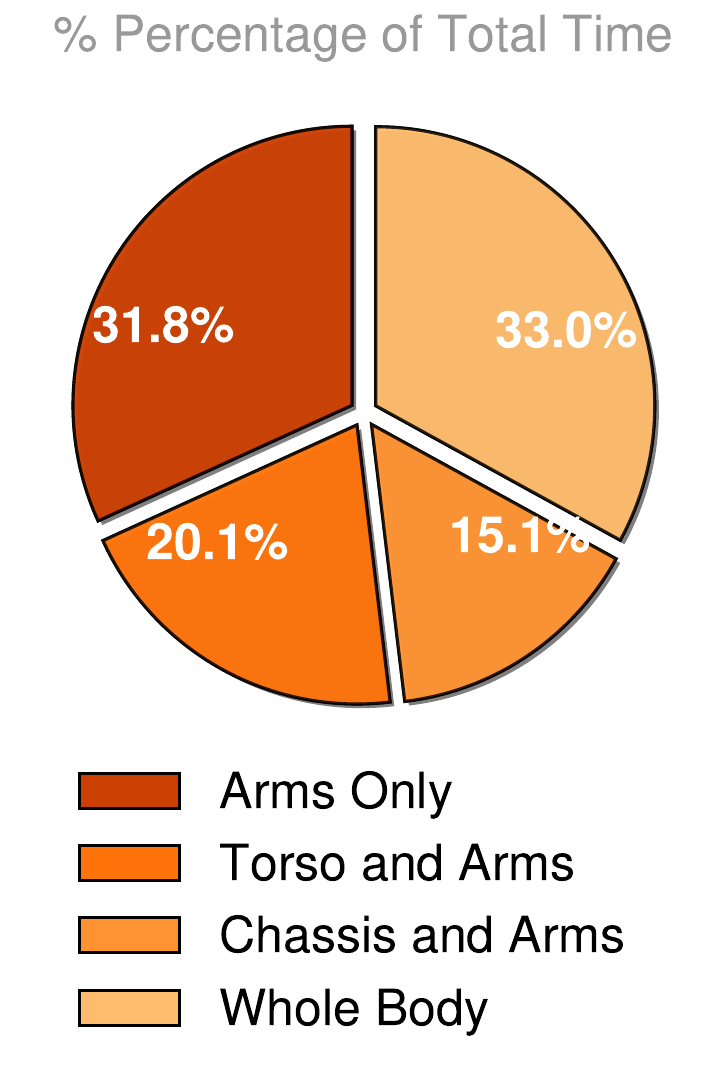}
        \caption{Body part usage.}
        \label{fig:body_part_usage} 
    \end{subfigure}
    \hfill
    \begin{subfigure}[b]{0.8\textwidth}
        \centering
        \includegraphics[width=\textwidth]{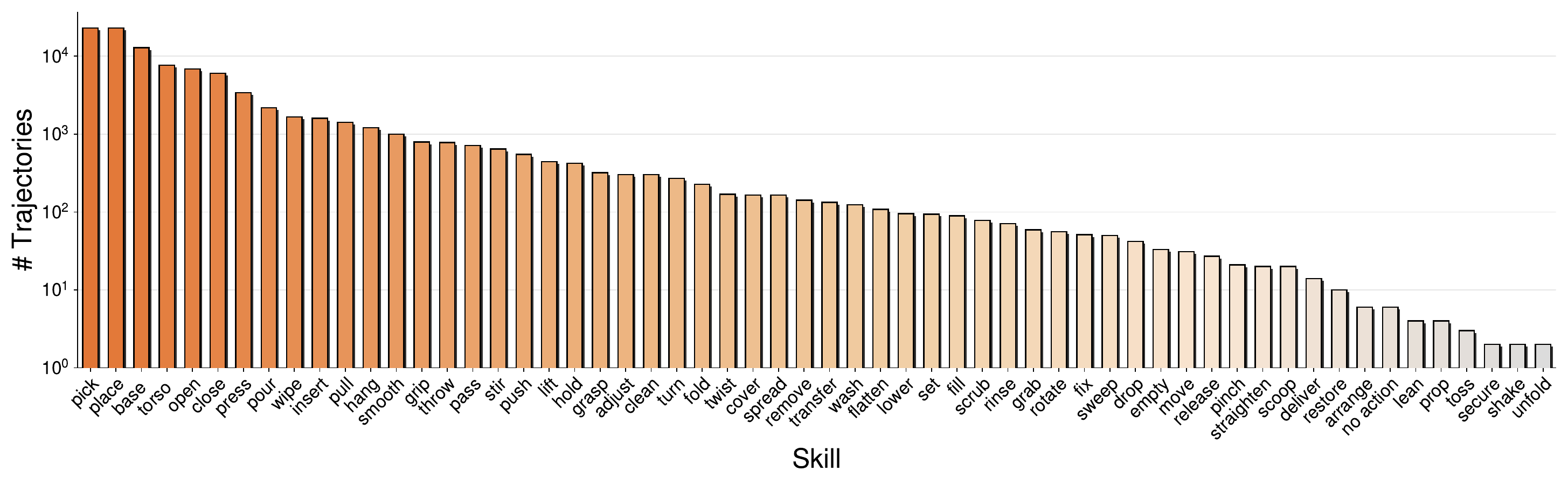}
        \caption{Skill distribution.}
        \label{fig:skill_distribution} 
    \end{subfigure}
    
    \caption{\textbf{Embodied behavior statistics.}  \textbf{(a)} A breakdown of interaction time by body part usage illustrates the variety of motions, from simple `Arms Only` manipulations to coordinated `Whole Body` movements. \textbf{(b)} The long-tail distribution of skills highlights the dataset's rich action vocabulary, covering both frequent, fundamental actions (e.g., `pick', `place') and a wide array of more specialized skills.}
    \label{fig:behavior_stats}
\end{figure}

\begin{figure}[tbp]
    \centering
    \includegraphics[width=0.9\linewidth]{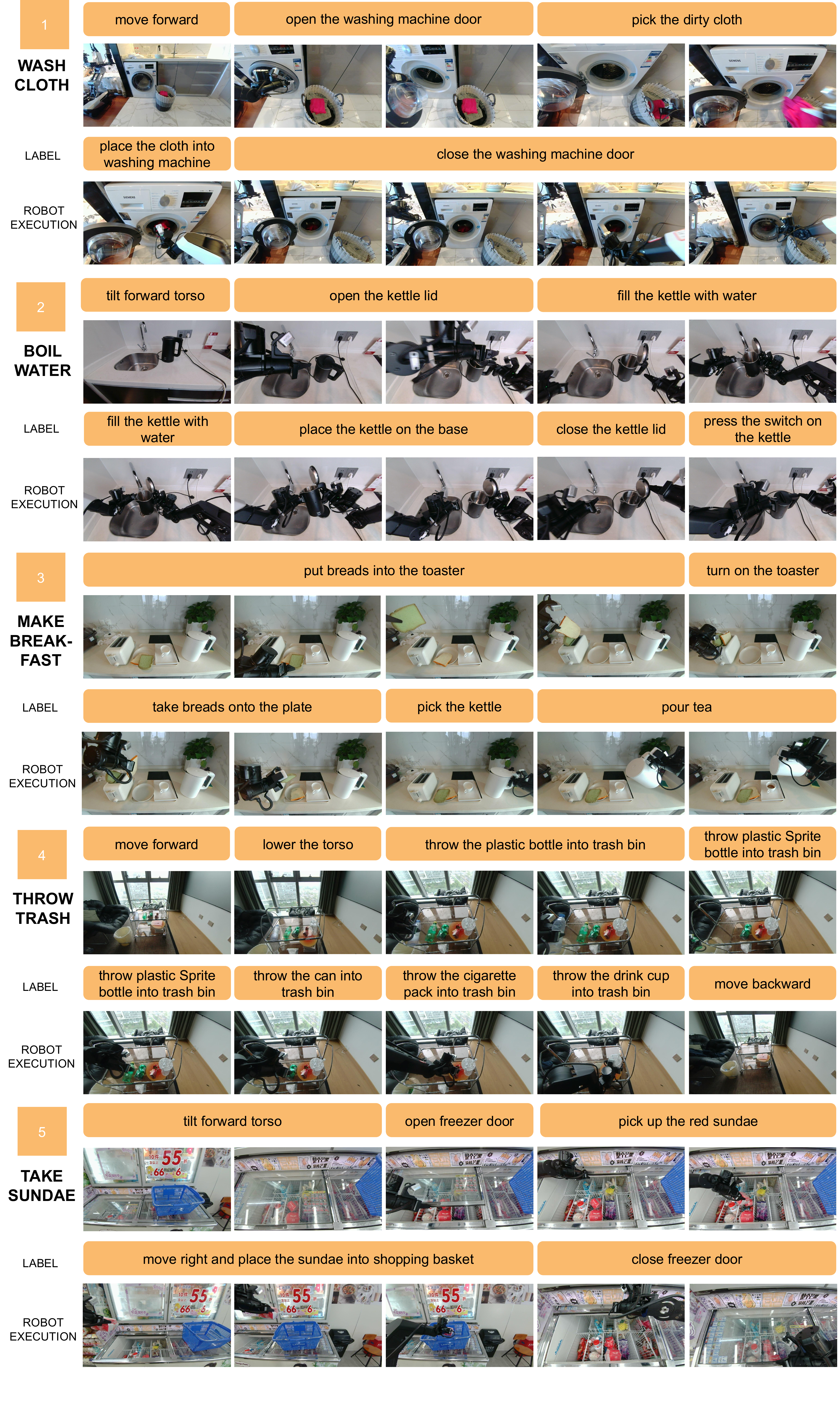}
    \caption{\textbf{Data samples with temporal subtask annotations.} Our dataset covers most scenes in people's daily lives, ranging from dual-arm manipulation to whole-body control robotics data, and provides high-quality, fine-grained subtask annotations.}
    \label{fig:dataset-teaser}
\end{figure}

\paragraph{Collection Guidelines.}
Data collection follows three guiding principles:  
(1) \textit{Observability} — visual streams contain all task-relevant cues, with key objects kept in view during manipulation;  
(2) \textit{Data quantity and quality} — for simple tasks, $\sim$100 well-executed demonstrations are sufficient; for more complex sequences, collection proceeds in a quality-first pilot phase before scaling;  
(3) \textit{Linguistic grounding} — each demonstration is annotated at the subtask level with structured language descriptions, enabling multi-modal alignment for VLA model training.

\paragraph{Environmental and Object Diversity.}
The dataset is collected at 11 physical sites—covering residential, catering, retail, and office spaces. 
Each site provides multiple operational zones, yielding a total of \textbf{50 unique scenes}. 
Object sets are sourced from real-world retail suppliers to ensure realistic visual and physical properties. 
For items that are unsafe or impractical to manipulate repeatedly (e.g., food), high-fidelity replicas are used to preserve visual realism while maintaining hygiene and efficiency.

\paragraph{Annotation Process.}
Each episode of data is segmented into atomic subtasks. 
Annotations follow a fixed schema, allowing annotators to select standardized descriptions rather than provide free-form text, thus improving labeling speed and consistency. 
Every segment undergoes rigorous quality checks at both the episode and clip levels; unqualified data—such as those containing operator mistakes or abnormal ROS topic frequencies—are excluded from training sets.

\paragraph{Comparison with Existing Datasets.}
We compare the Galaxea Open-World Dataset with prior large-scale robot datasets such as BridgeData~\cite{ebert2021bridge}, RT-1~\cite{zitkovich2023rt}, Open-X-Embodiment~\cite{vuong2023open}, and AgiBot World~\cite{bu2025agibot}. 
Our dataset offers (i) \emph{single-embodiment consistency} across a wide range of skills and environments, (ii) fine-grained subtask-level annotations for precise multi-modal alignment, and (iii) significantly higher scene diversity in real-world settings.
These properties make it a strong benchmark for studying generalizable VLA models that can operate reliably in unstructured human environments.

\section{Method}

\subsection{Dual System Overview}
As shown in Figure~\ref{fig:teaser}, our G0 dual system is comprised of a fast-response System-1 and a deliberative System-2. System-1 is a VLA model responsible for perceiving the environment, interpreting subtask instructions, and executing actions. It is an end-to-end vision-language-action (VLA) model designed to control a bi-manual robot with a mobile torso and a chassis. At each time step $t$, it generates an action chunk $\boldsymbol{A}_t=a_{t:t+k}$ with horizon $k$ conditioned on the input language instructio\textbf{}n $l$, visual observations from three cameras $o_t$, and robot's proprioceptive state $s_t$. G0-VLA first embeds the robot's visual observations and language instructions with the pre-trained VLM, then generates continuous action with an flow matching action expert conditioned on VLM's KV cache. 
Meanwhile, System-2 operates as a VLM that processes high-level natural language task instructions, understands the scene, and subsequently plans subtask instructions for System-1.

We adopt different training recipes for these two models. The G0-VLM is trained with the image and subtask annotation pairs derived from our Galaxea Open-World Dataset.
Regarding the G0-VLA model, we introduce a 3-stage training strategy that leverages diverse datasets to progressively refine its performance and adaptability. 
Specifically, \textit{pre-training Stage-1} aims to acquire a general prior model from extensive and diverse robotics data, combining Galaxea's proprietary large-scale unlabeled dataset with additional publicly available robotics datasets. \textit{pre-training Stage-2} subsequently specializes the model to capture the configuration, dynamics, and kinematics of a single embodiment through diverse physical interactions recorded within our high-quality data. Finally, the \textit{Post-training} stage fine-tunes the model to master a targeted set of tasks using a small number of high-quality demonstration data. This section elaborates on each stage in detail.

\subsection{G0-VLA Pre-training Stage-1}
In the stage 1 pretraining, we only train the VLM component. To enable the VLM’s language model backbone to predict robot actions, we adopt FAST tokenizer as our action tokenizer, which converts raw continuous action chunks into a sequence of discrete indices. In this way, we can train VLM with a standard cross-entropy loss to predict the next action tokens in an autoregressive fashion. Specifically, given the image observations $o_t$, language instruction $l_t$, and proprioceptive state $s_t$ at time $t$ the policy is trained to model the conditional distribution over action tokens:
$$ p(\mathbf{A}^{d}_t) = \prod_{i=1}^{N} p(a^{d}_i \mid a^{d}_{<i}, o_t, l_t, s_t), $$
where $\mathbf{A}^{d}_t$ denotes the $N$ discrete action tokens $a^{d}$ produced by the action tokenizer. The VLM is initialized from PaLiGemma~\cite{beyer2024paligemma} and consists of a SigLIP vision encoder, followed by a single-layer MLP projector and a standard Transformer. The vision encoder and projector together transform the three input images into a one-dimensional sequence of embeddings, which then attend to the tokenized language instruction, proprioceptive state, and previously predicted action tokens via attention within the Transformer.

For the data used in Stage-1 Pretraining, we train the VLM on a diverse mixture of robot embodiment recordings. This includes approximately 1,000 hours of OXE trajectories, 500 hours from the GalaXea Open-World Dataset (where only the high-level task descriptions are used, excluding low-level language annotations), and 200 hours of in-house data containing high-level task descriptions only.

The motivation for training only the VLM in this stage can be explained from two perspectives: (1) Training data is collected from various embodiments, and the quality of annotations and the accuracy of their corresponding actions are inconsistent. Therefore, the action expert cannot learn sufficiently informative knowledge from these data. (2) Diffusion loss may harm the learning process if applied before the model has converged to generate stable representations.

\begin{figure}[tbp]
        \centering
        \hspace*{-0.025\textwidth}  
\includegraphics[width=1.05\textwidth]{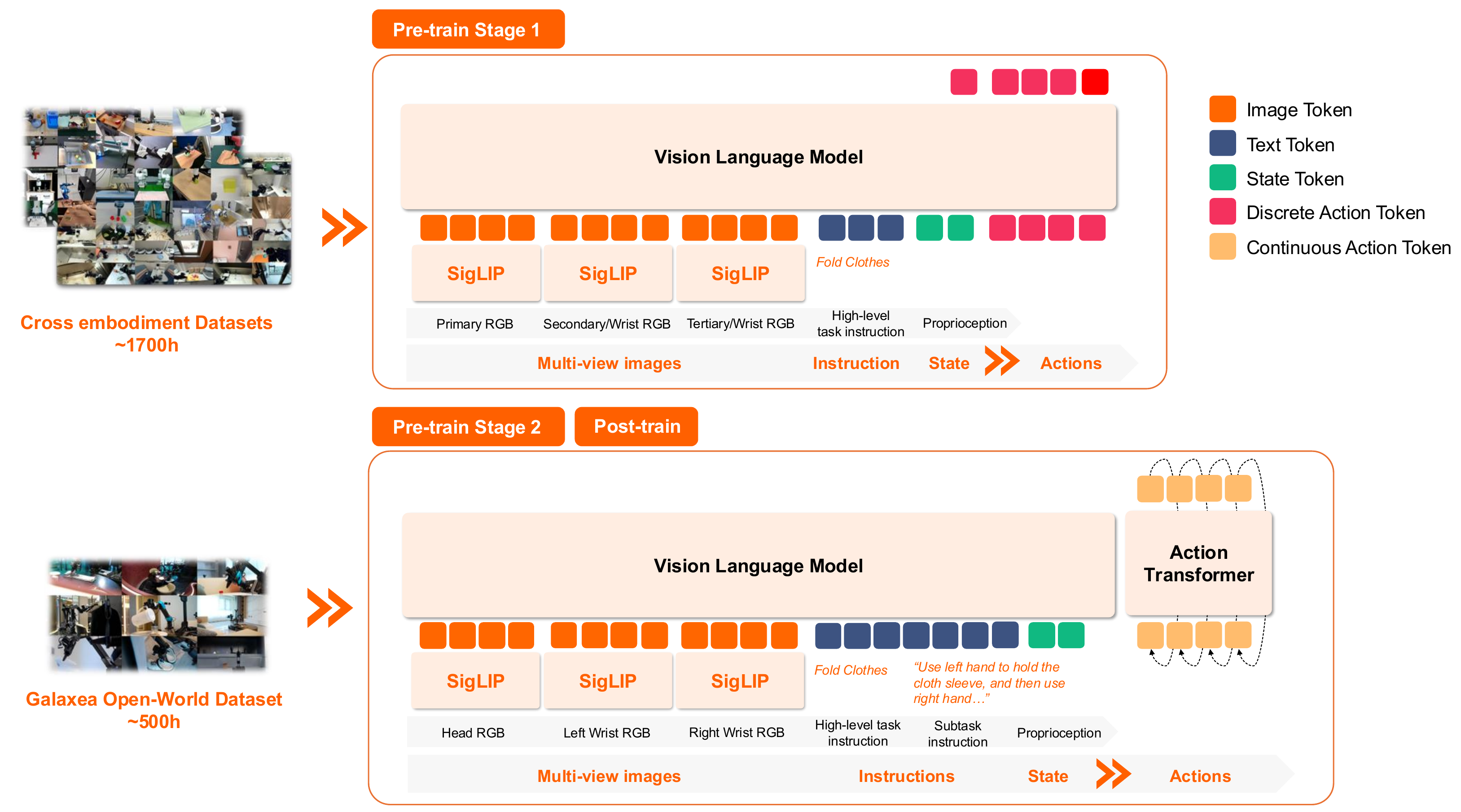} 
    \caption{G0-VLA architecture and our 3-stage training pipeline. Stage 1 pre-trains a vision-language model on cross-embodiment data in an autoregressive manner. Stage 2 and post-train share the same model structure, trained on Galaxea open-world data with embodiment-specific views and high-level and subtask instructions, by supervising the Action Transformer’s action reconstruction with a flow-matching loss. Color codes indicate token modalities.}
    \label{fig:vla_model}
\end{figure}

\subsection{G0-VLA Pre-training Stage-2}
In this stage, we train our System-1 VLA on our labeled Galaxea Open-World Dataset. The VLA consists of the pre-trained VLM and a newly initialized action expert. The action expert generates continuous actions conditioned on proprioceptive states and the representations generated by the VLM.
Specifically, we train our VLA by maximizing the following objective:
$$
\max_{\theta} \; \mathbb{E}_{p(\boldsymbol{A}_{t}, {o}_t, l_t, {s}_t)} \left[ \log \pi_{\theta}(\boldsymbol{A}_{t} \mid {o}_t, l_t, {s}_t) \right]
$$
with a flow-matching loss
$$
\mathcal{L}_{\text{flow}}(\theta) = \mathbb{E}_{p(\boldsymbol{A}_t^\tau \mid {o}_t, l_t, {s}_t)} \left[
\left\|
v_{\theta}(\boldsymbol{A}_t^\tau, \tau, {o}_t, l_t, {s}_t)
-
\boldsymbol{u}(\bm{A}_t^\tau \mid \boldsymbol{A}_t)
\right\|^2
\right]
$$
Here, $\boldsymbol{A}_t$ denotes the action chunk from time $t$ with horizon $H$, ${o}_t$ is the visual observation, $l_t$ is the language instruction, and ${s}_t$ is the proprioceptive state. $\boldsymbol{A}_t^\tau$ is the interpolated noisy action $\boldsymbol{A}_t^\tau=\tau\boldsymbol{A}_t+(1-\tau)\epsilon$. $v_{\theta}(\cdot)$ is the flow predicted by the VLA and $\boldsymbol{u}(\cdot)$ is the target flow derived from the action trajectory.

Pre-training Stage-2 focuses on improving the action precision and language grounding capabilities, enabled by two key features of the Galaxea Open-World Dataset: 1) \textbf{single embodiment}: All trajectories are collected on a single robotic platform, ensuring a consistent action space and eliminating the need for the action expert to adapt across embodiments. 2) \textbf{language-action alignment}: Instructions and trajectories are segmented at the subtask level, producing fine-grained language-action pairs. This promotes a stronger correspondence between instructions and robot actions.


\subsection{VLA Post-training: Task-oriented Training}
To test the generalization ability of pre-trained models, we fine-tune our VLA with different pre-trained weights on downstream tasks, using the same training objective as Stage-2. For each task, we limit the fine-tuning data to a maximum of 100 trajectories.

\subsection{G0-VLM Training}

G0-VLM is the high-level planner of the dual system which serves several purposes: it interprets humans' high-level instructions, responds with verbal language, performs task planning, and sends low-level atomic action instructions to the G0-VLA for execution. In our implementation, we adopt the open-source Qwen2.5-VL~\cite{Qwen2.5-VL} as a starting point and conduct instruction tuning using data sampled from the Galaxea Open-World dataset. 

To train G0-VLM in a scalable manner, we utilize human-annotated subtasks alongside synthesized human‑style high‑level instructions. 
First, we sample episodes from the Galaxea Open-World dataset. While sampling, key frames, defined as moments when the subtask is close to terminating, or gripper state changes, are assigned higher sampling weights to facilitate learning of task transitions.
Then, we extract head camera images and subtask annotations. In order for the VLM to handle task planning over long temporal contexts, we as well feed the model with $k$-frame historic image observations and robot actions from preceding time steps at 1-second intervals. The resulting dataset $D_\text{labeled}$ contains task name, robot observations $o_{t-k}, ..., o_t$, and subtask instructions $l_{t-k}, ..., l_t$. 

Subsequently, we apply a reasoning LLM (DeepSeek-R1) on $D_\text{labeled}$ to produce human‑style high‑level instructions and the robot's response to humans. We prompt the LLM with each episode’s task name (e.g., \textit{pull and push chairs}), historic and current subtasks, and the next subtask; the LLM then reasons over the prompt, analyzes the entire action sequence, imagining the real-world scenario, and ultimately produces a human‑style verbal instruction (e.g., \textit{"I am going to be seated, could you help pull the chair out?"}) and the robot's verbal response to the human (e.g., \textit{"I am working on it!"}).
Here, we do not feed image observations to the LLM, as we argue that the reasoning capabilities of an LLM are sufficient to infer task scenarios from our high-quality atomic action annotations. 


\section{Evaluating G0-VLA}
\begin{figure}[tb]
        \centering
\includegraphics[width=1.0\textwidth]{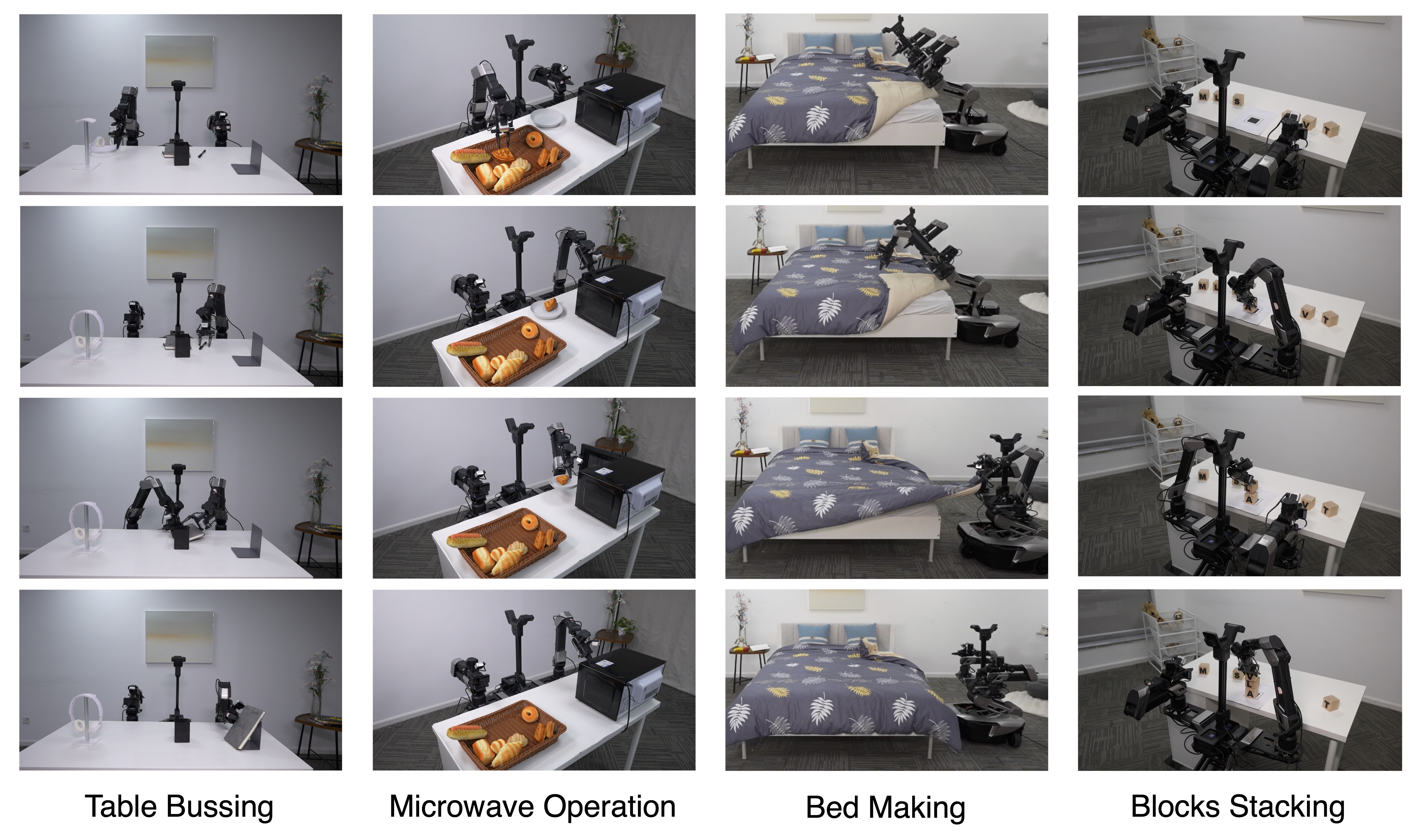} 
    \caption{Evaluation benchmarks.}
    \label{fig:exp-setup}
\end{figure}

In this section, we build challenging benchmarks and conduct fine-tuning experiments to evaluate G0-VLA model and the effectiveness of our proposed dataset. Each task is carefully designed to probe a specific capability of our model. At the heart of our inquiry lies a central question: \textbf{How does pre-training data influence VLA?} We explore this through three perspectives:
(1) Does pre-training enhance fine-tuning performance on downstream tasks? How much does the pre-trained weights matter?
(2) Can pre-training on a single embodiment accelerate few-shot transfer?
(3) How do single-embodiment and cross-embodiment pre-training compare in embodiment-specific actions?

Our benchmarks consist of the following tasks:

\textbf{Table bussing}: The robot is required to organize a cluttered desk by placing pens into a pen holder, picking up and hanging headphones, and moving a book onto a book stand. This task evaluates the model’s capability in precise pick-and-place, coordinated dual-arm manipulation, and maintaining object stability.

\textbf{Microwave operation}: The robot opens a microwave door, places food onto a plate, transfers the plate into the microwave, and then closes the door to initiate heating. This task assesses the model’s ability in interacting with household appliances and executing multi-step manipulation sequences.

\textbf{Bed Making}: The robot is asked to tidy up the messy quilt on the bed to make the quilt flat and neat. This task emphasizes whole-body control, requiring coordination of the chassis, torso, and arms for effective execution.

\textbf{Blocks Stacking}: The robot is asked to build the blocks to form specific words. This task tests the model's ability in language following and precise pick-and-place.

We evaluate these benchmarks by progress score, where the details of progress for each task are defined in the appendix. For reproducibility, we run each test 10 times and get the average score of each task. 

\subsection{Pre-trained Weights}
\begin{figure}[t]
    \centering
\includegraphics[width=1.0\textwidth]{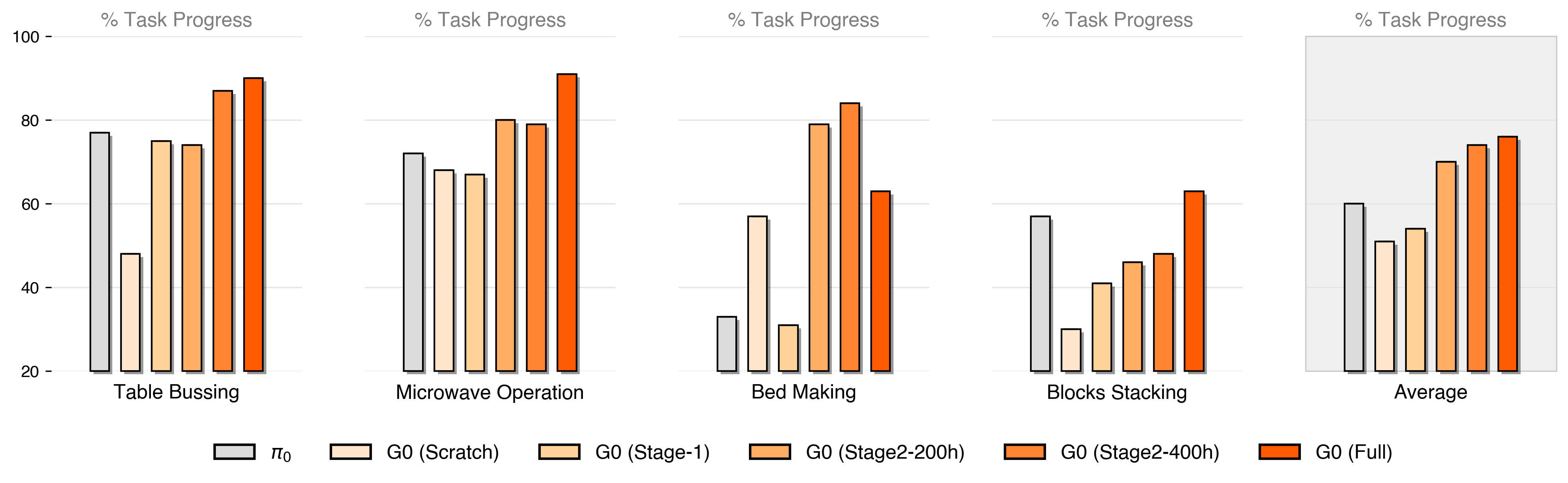} 
    \caption{\textbf{Fine-tuning benchmark results of different pre-trained VLAs.} G0 (Full) achieves the highest average progress score, excelling in object-picking tasks such as \textbf{Table Bussing}, \textbf{Microwave Operation}, and \textbf{Bed Making}. G0 (Stage-2) lead in language following, action consistency, and whole-body control. G0 (Stage-1) performs the worst among pre-trained models, highlighting the necessity of uniform-embodiment pre-training.}
    \label{fig:table_1}
\end{figure}

In this experiment, we test the effectiveness of different pre-trained weights. We fine-tune the pre-trained models on our proposed benchmarks, using 100 training trajectories per task (each ranging from 30 seconds to 1 minute in duration). The following configurations are evaluated:

\begin{itemize}
    \item G0 (Stage-1): VLA with only Stage-1 pre-training.
    \item G0 (Stage-2 200h): VLA with only Stage-2 pre-training (200 hours of data).
    \item G0 (Stage-2 400h): VLA with only Stage-2 pre-training (400 hours of data).
    \item G0 (Full): VLA with Stage-1 followed by Stage-2 pre-training (400 hours of data).
    \item G0 (Scratch): VLA without any action pre-training (initialized from the original VLM weights).
    \item \(\pi_0\): \(\pi_0\)~\cite{black2024pi_0} with officially released pre-trained weights as a baseline.
\end{itemize}

All models are fine-tuned under identical settings for four epochs. The results are shown in Figure~\ref{fig:table_1}. Overall, G0 (Full) achieves the highest average progress score. In particular, it demonstrates superior object-picking ability in \textbf{Table Bussing}, \textbf{Microwave Operation}, and \textbf{Bed Making}. G0 (Stage-2 400h) and G0 (Stage-2 200h) achieve the best performance in language following and action consistency, as well as the strongest whole-body control capabilities, which are further discussed in Section~\ref{sec:embodiment_specific_actions}. By contrast, G0 (Stage-1) performs the worst among all pre-trained models, underscoring the importance of single-embodiment pre-training.

We observe that Stage-1 pre-training primarily enhances VLA’s ability to perform simple and universal action patterns, such as pick-and-place and push-and-pull. Meanwhile, Stage-2 pre-training grounds the model specifically to our robot platform, leading to improved action stability and instruction following.

\subsection{Few-shot Transfer}
\begin{figure}[htbp]
        \centering
\includegraphics[width=0.9\textwidth]{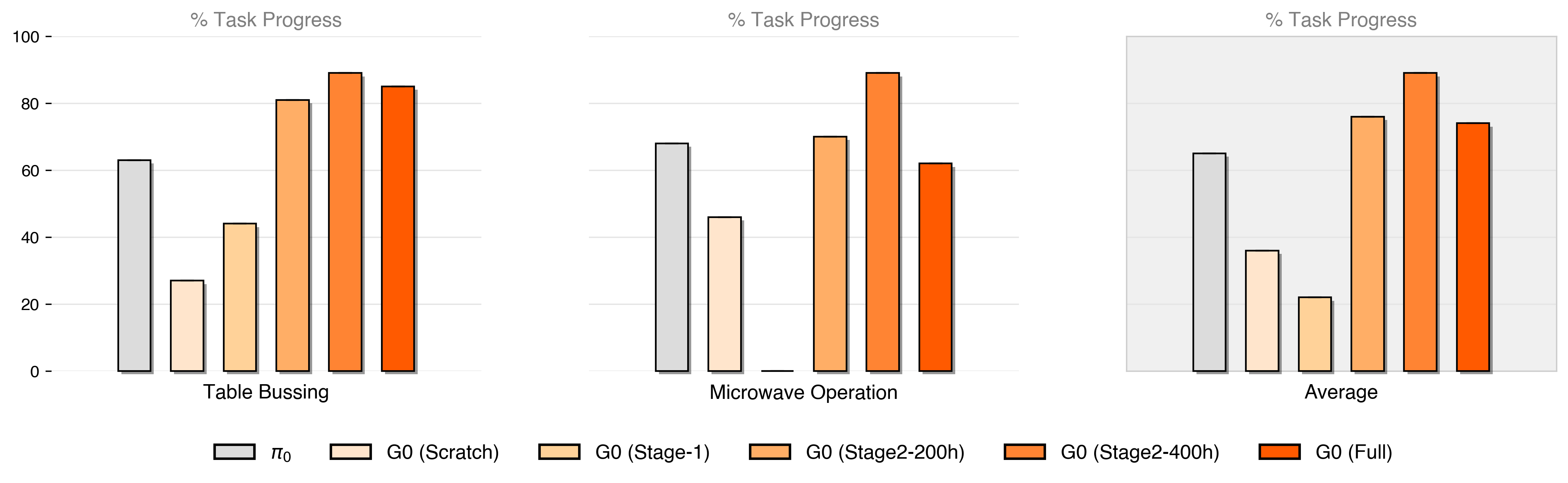} 
    \caption{\textbf{Few-shot performance of VLAs.} Few-shot transfer performance on \textbf{Table Bussing} and \textbf{Microwave Operation}. Stage-2 pre-training markedly improves success rates and execution smoothness, while Stage-1 pre-training alone offers no clear advantage over training from scratch.}
    \label{fig:fewshot}
\end{figure}

In this part, we specifically assess the few-shot transfer capability of our VLA. We fine-tune the model using only 20 trajectories for each of two tasks: \textbf{Table Bussing} and \textbf{Microwave Operation}. Each model is fine-tuned with the same setting for 10 epochs. 

As shown in Figure~\ref{fig:fewshot}, models with Stage-2 pre-training significantly outperform those without. 
In addition to this quantitative improvement, we also observe that these models produce noticeably smoother and more stable actions during execution. These results suggest that single-embodiment pre-training substantially enhances few-shot generalization within the same embodiment, underscoring the importance of single-embodiment data in our Galaxea Open-World Dataset.
Notably, models pre-trained solely with Stage-1 do not show a clear advantage over models trained from scratch. This indicates that cross-embodiment action pre-training alone may be insufficient for a model's ability to quickly adapt to a new embodiment in few-shot settings.

\subsection{Embodiment-specific Actions}
\label{sec:embodiment_specific_actions}
\begin{figure}[htbp]
        \centering
\includegraphics[width=1.0\textwidth]{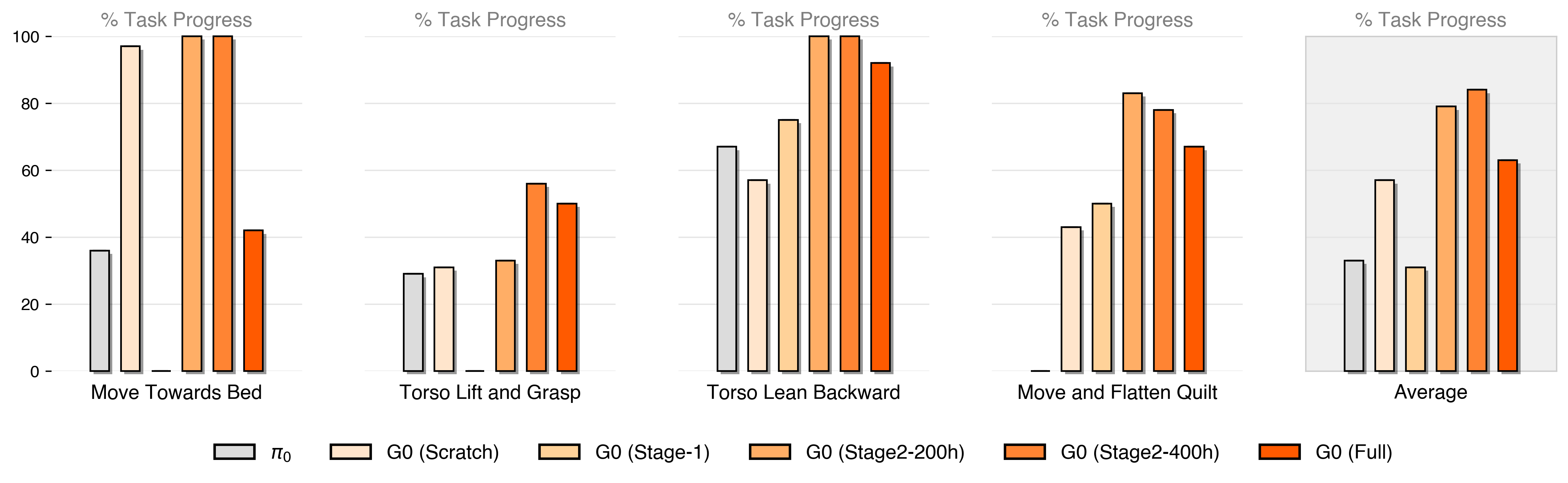} 
    \caption{\textbf{Per-skill progress scores on the Bed Making task.} Stage-2 single-embodiment pre-training substantially improves chassis, torso control, while cross-embodiment pre-training (Stage-1, $\pi_0$) yields weaker performance, in some cases worse than training from scratch.}
    \label{fig:embodiment_specific_action}
\end{figure}

In this section, we present a detailed analysis of embodiment-specific actions. The \textbf{Bed Making} is a long-horizon task which requires frequent, coordinated, and precise whole-body control, including the chassis, torso and arms. These are embodiment-specific behaviors that are not represented in cross-embodiment datasets such as OXE. We report the progress scores by skill in Figure~\ref{fig:embodiment_specific_action}.

Stage-2 pre-training on single-embodiment data significantly improves the model’s performance on these embodiment-specific skills, suggesting that such capabilities are effectively acquired during this stage of pre-training. In contrast, models trained with cross-embodiment data (e.g., Stage-1 pre-training and $\pi_0$) demonstrate substantially weaker instruction following for chassis-related actions and less accurate torso control. In some cases, they even underperform compared to models trained from scratch. 
We hypothesize that the large embodiment gap between our robot and those in the OXE dataset used for Stage-1 pre-training hinders the model’s ability to acquire embodiment-specific skills. These findings highlight the need to carefully design the use of cross-embodiment data in pre-training strategies to ensure positive knowledge transfer.

\section{Evaluating G0-VLM}



In our dual-system framework, G0-VLM serves as the task planner, processing human instructions and environmental observations to generate executable commands for the downstream VLA module. The effectiveness of this process depends critically on two aspects: the accuracy of command-observation alignment, which ensures the VLM correctly interprets perceptual inputs, and the fidelity of action primitives, which determines whether the VLA can properly execute the generated commands. We therefore design our evaluation metrics to rigorously assess both aspects of VLM performance.

We investigate two key questions: (i) whether fine-tuning is necessary compared to using pretrained models directly, and (ii) how supervised fine-tuning (SFT) can enhance the VLM's capabilities in robotic tasks, particularly in improving action-grounding accuracy. To address these, we benchmark several established models, including Gemini-2.5-pro and Qwen2.5-VL (with variants of different parameter sizes), against our fine-tuned versions. To ensure a fair comparison, we design standardized prompts containing task-specific instructions, atomic action options, and output examples for all baseline models. This setup restricts the models to selecting only from the provided options, eliminating variability due to prompt design and focusing evaluation on their core capabilities.

Table \ref{tab:vlm} demonstrates that our fine-tuned model surpasses baseline accuracy by over $50\%$, with task-specific tuning enabling language instructions that are directly executable by VLAs. This validates our key hypothesis that robotic applications require not just general-purpose vision-language understanding, but precisely aligned action primitives through domain adaptation.

\begin{table}[h]
\caption{Instruction accuracy in benchmark tasks~(\%).}
\label{tab:vlm}
    \centering
    \begin{tabular}{c|cccc}
        \toprule
         Model & Table bussing & Microwave operation & Make the bed & Build blocks\\
         \midrule
         Gemini-2.5-pro & 32.0 & 15.8 & 54.2 & 55.0 \\
         Qwen2.5-VL-72B & 26.3 & 16.8 & 48.1 & 21.7 \\
         Qwen2.5-VL-32B & 21.3 & 14.8 & 54.2 & 21.0 \\
         Qwen2.5-VL-7B & 26.3 & 17.2 & 46.9 & 24.7 \\
         \midrule
         G0-VLM & 83.3 & 74.2 & 78.2 & 75.6 \\ 
        \bottomrule
    \end{tabular}
\end{table}

\section{Conclusion}

We have introduced Galaxea Open-World Dataset, a large-scale, high-fidelity, and richly annotated resource designed to accelerate research in robotic mobile manipulation. 
By pre-training on the dataset, we present G0, a dual system composed of a VLM for planning and a VLA model for execution. G0 achieves state-of-the-art performance across a diverse set of benchmarks.

\section{Contributors}
\noindent\textbf{Dataset operation.} Tao Jiang, Jianning Cui, Xiao Liu

\noindent\textbf{Policy training and evaluation.} Tianyuan Yuan, Yicheng Liu, Chenhao Lu, Shuiqi Cheng, Jianning Cui, Xiao Liu, Tao Jiang

\noindent\textbf{Project supervision.} Hang Zhao, Huazhe Xu, Jiyang Gao

\noindent We sincerely thank the data collection team, the hardware team and the marketing team at Galaxea for their support.
 
\newpage
\bibliographystyle{unsrtnat}
\bibliography{galaxea}


\appendix

\section{Appendix / supplemental material}

\subsection{Evaluation Details}
Here we present the score rubric that measures the progress of each task which is presented as follows:

\textbf{Table Bussing}: This task is scored out of 6 points. Each action is treated as a pick-and-place operation, with 1 point awarded for a successful pick and 1 point for a successful place.

\textbf{Microwave Operation}: This task is scored out of 5 points, awarded for the following steps: picking the correct food, placing it on the plate, opening the microwave, placing the plate inside, and closing the microwave.

\textbf{Bed Making}: This task is scored out of 4 points, corresponding to: moving toward the bed, lifting the torso and grasping the quilt, leaning the torso back, and moving to flatten the quilt.

\textbf{Blocks Stacking}: This task is scored out of 6 points. Each action is treated as a pick-and-stack operation, with 1 point awarded for a successful pick and 1 point for a successful stack.

\end{document}